\documentclass[10pt,twocolumn,letterpaper]{article}

\usepackage{cvpr}
\usepackage{times}
\usepackage{epsfig}
\usepackage{graphicx}
\usepackage{amsmath}
\usepackage{amssymb}
\usepackage{multirow}
\def\cm{\checkmark}
\usepackage{kotex}
\usepackage{color}
\usepackage{footnote}

\newcommand{\customfootnotetext}[2]{
    {\renewcommand{\thefootnote}{#1}\footnotetext[0]{#2}}
}

\usepackage[pagebackref=true,breaklinks=true,letterpaper=true,colorlinks,bookmarks=false]{hyperref}

\definecolor{cadmiumgreen}{rgb}{0.0, 0.42, 0.24}

\newcommand{\Fref}[1]{Figure \ref{#1}}
\newcommand{\Sref}[1]{Section \ref{#1}}
\newcommand{\Tref}[1]{Table \ref{#1}}

\cvprfinalcopy 

\def\cvprPaperID{95} 

\begin{document}

\title{SimUSR: A Simple but Strong Baseline for Unsupervised Image Super-resolution}

\author{
Namhyuk Ahn\textsuperscript{$\dagger$}\footnotemark[1]\\
Ajou University\\
{\tt\small aa0dfg@ajou.ac.kr}
\and
Jaejun Yoo\footnotemark[1]\\
EPFL\\
{\tt\small jaejun.yoo88@gmail.com}
\and
Kyung-Ah Sohn\textsuperscript{$\ddagger$}\\
Ajou University\\
{\tt\small kasohn@ajou.ac.kr}
}

\maketitle

\begin{abstract}
In this paper, we tackle a fully unsupervised super-resolution problem, \textit{i.e.,} neither paired images nor ground truth HR images. We assume that low resolution (LR) images are relatively easy to collect compared to high resolution (HR) images. By allowing multiple LR images, we build a set of pseudo pairs by denoising and downsampling LR images and cast the original unsupervised problem into a supervised learning problem but in one level lower. 
Though this line of study is easy to think of and thus should have been investigated prior to any complicated unsupervised methods, surprisingly, there are currently none. Even more, 
we show that this simple method outperforms the state-of-the-art unsupervised method with a dramatically shorter latency at runtime, 
and significantly reduces the gap to the HR supervised models. 
We submitted our method in NTIRE 2020 super-resolution challenge and won 1st in PSNR, 2nd in SSIM, and 13th in LPIPS. 
This simple method should be used as the baseline to beat in the future, especially when multiple LR images are allowed during the training phase. However, even in the zero-shot condition, we argue that this method can serve as a useful baseline to see the gap between supervised and unsupervised frameworks. 
\end{abstract}

\section{Introduction}
\customfootnotetext{$\dagger$}{~Most work was done while in NAVER Corp.}
\customfootnotetext{*}{~indicates equal contribution. $\ddagger$ indicates corresponding author.}

\begin{figure}[!t]
\centering
\includegraphics[width=0.24\linewidth]{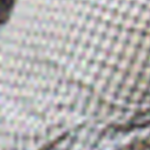}
\includegraphics[width=0.24\linewidth]{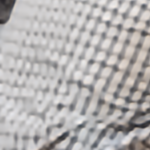}
\includegraphics[width=0.24\linewidth]{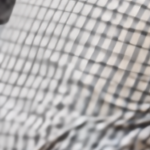}
\includegraphics[width=0.24\linewidth]{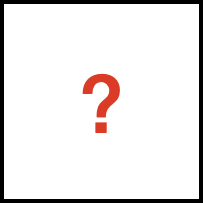}

\includegraphics[width=0.24\linewidth]{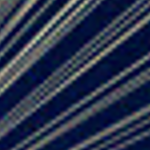}
\includegraphics[width=0.24\linewidth]{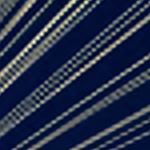}
\includegraphics[width=0.24\linewidth]{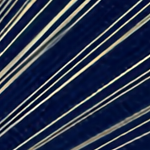}
\includegraphics[width=0.24\linewidth]{./assets/teaser/HR.png}

\includegraphics[width=0.24\linewidth]{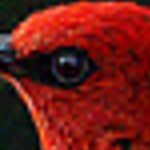}
\includegraphics[width=0.24\linewidth]{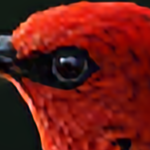}
\includegraphics[width=0.24\linewidth]{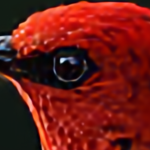}
\includegraphics[width=0.24\linewidth]{./assets/teaser/HR.png}

\includegraphics[width=0.24\linewidth]{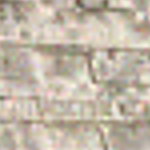}
\includegraphics[width=0.24\linewidth]{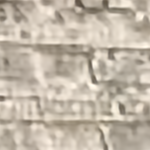}
\includegraphics[width=0.24\linewidth]{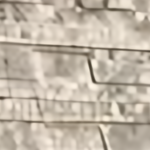}
\includegraphics[width=0.24\linewidth]{./assets/teaser/HR.png}
\makebox[0.24\linewidth][c]{\small{Original}}
\makebox[0.24\linewidth][c]{\small{ZSSR+BM3D}}
\makebox[0.24\linewidth][c]{\small{Ours+BM3D}}
\makebox[0.24\linewidth][c]{\small{Ground Truth}}
\vspace{1em}
\caption{NTIRE 2020 Super-resolution challenge ($\times$4)~\cite{NTIRE2020RWSRchallenge} images (1st column).  ZSSR~\cite{shocher2018zero} with BM3D~\cite{dabov2007image} results (2nd column). Our SimUSR with BM3D results (3rd column). Ground truth HR image (4th column) is not available in this setup. Our method achieves superior SR performance for all the cases.} 
\label{fig:teaser}
\end{figure}
Single image super-resolution (SISR) is a longstanding task in computer vision area,  
which focuses on recovering a high-resolution (HR) image from a single low-resolution (LR) image.
Since this task has to solve a one-to-many mapping problem, building an effective SISR method is challenging.
Despite the difficulties, thanks to the developments of deep learning and large-scale datasets with high quality images, 
many learning-based SR methods~\cite{ahn2018fast,dong2015image,kim2016accurate,ledig2017photo,lim2017enhanced,zhang2018image} have recently shown prominent performance gains over traditional optimization-based approaches~\cite{timofte2014a+}. With the power of the deep neural network, they enjoy the dramatic boost of the SR performance by stacking a lot of layers~\cite{zhang2018image} or widen the network~\cite{lim2017enhanced}.

While most of the deep learning-based SR methods heavily rely on a large number of image pairs (supervised SR), unfortunately, such a large-scale and high quality dataset is not always accessible, especially when we deal with a real environment. 
A few recent works have proposed a workaround solving an unpaired SR task~\cite{fritsche2019frequency,lugmayr2019unsupervised,yuan2018unsupervised}.
Since this setup does not require full supervision of LR and HR image pairs but images from each domain, it is a more realistic scenario for many real-world applications.
However, in many applications (\eg, medical image), gathering HR (or clean) images itself requires a lot of efforts or sometimes even impossible.


To address this, Shocher~\etal~\cite{shocher2018zero} have proposed a fully unsupervised method, called zero-shot super-resolution (ZSSR), which performs both training and testing at runtime using only a single LR test image. By learning a mapping from a scale-down version of the LR image to itself, ZSSR learns to exploit internal image statistics to super-resolve the given image. It outperformed the previous internal SR methods~\cite{huang2015single} in a huge margin with a high flexibility because the model can easily be adapted to any unknown degradation or downsample kernel.

\begin{figure*}[t]
\centering
\includegraphics[width=\linewidth]{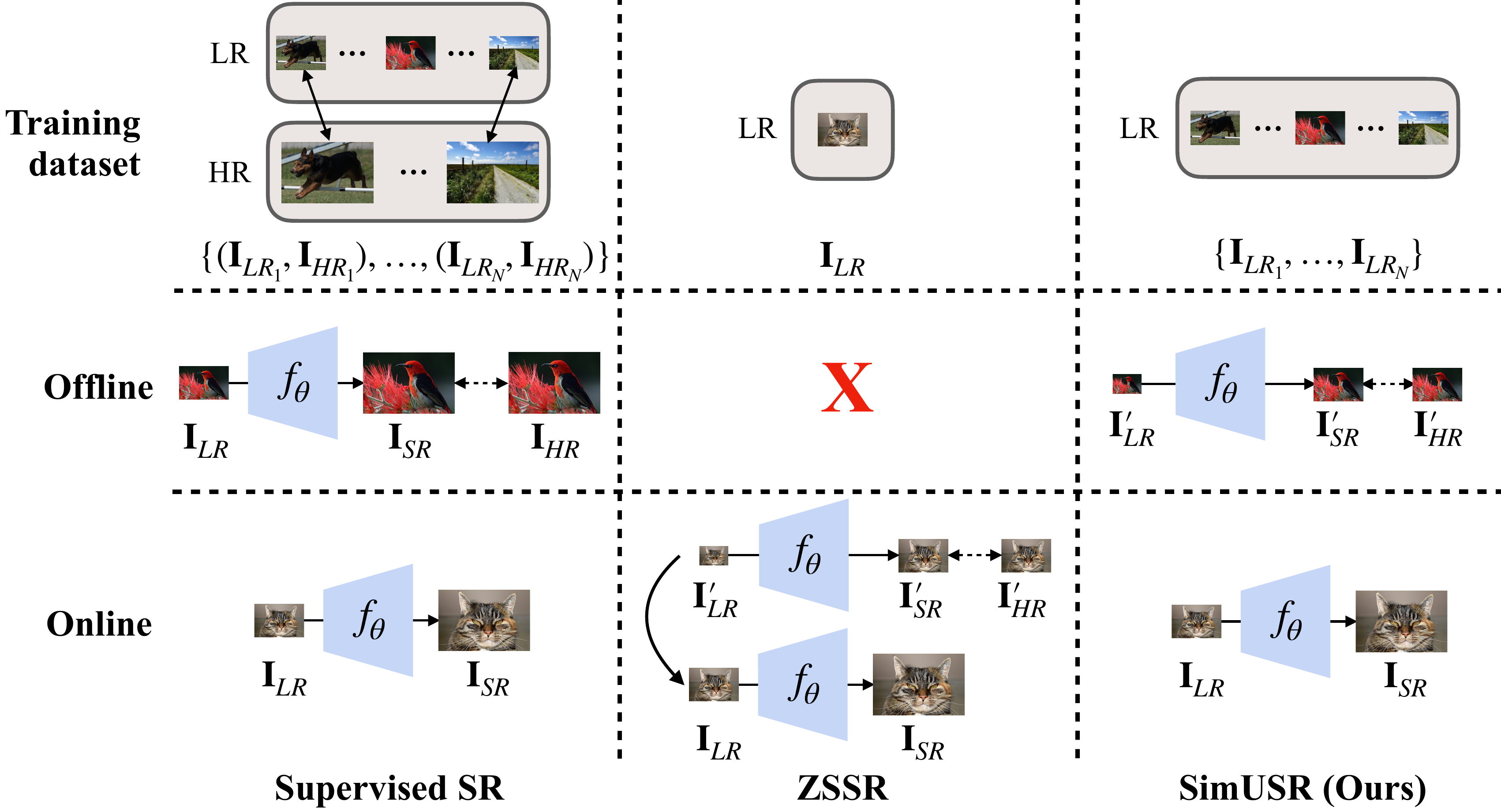}
\caption{Schematic comparison of the supervised SR, ZSSR~\cite{shocher2018zero}, and our SimUSR. We analyze current SR approaches in terms of the training dataset, offline phase, and online phase. The offline phase is operated beforehand the user's inference request (\textit{i.e.} training process of the supervised SR). Online phase denotes runtime. (\textbf{1st row}) While supervised SR requires the LR-HR pairs, ZSSR and SimUSR use LR images only, making them more applicable to the real-world SR scenarios. ZSSR utilizes only a single test LR image and performs both optimization and inference at runtime. (\textbf{2nd, 3rd rows}) On the other hand, SimUSR exploits additional LR images and follows a similar procedure to the supervised setup, where the model is first trained offline and inference is done online.}
\label{fig:overview}
\end{figure*}

However, ZSSR has several drawbacks. \textbf{1)} It requires an online optimization procedure at runtime. Since it needs at least 1K steps (both forward and backward propagation), the latency of the ZSSR is extremely high. \textbf{2)} It is difficult to benefit from a large capacity network. Because ZSSR has to perform online training on a single image, the model should be able to quickly adapt to the given image while avoiding the overfitting issue, which limits ZSSR to use a shallow network architecture. \textbf{3)} When noises are present in LR images, ZSSR shows deteriorated performance because the model can never learn to denoise, and even after adding a denoising module, it suffers from its restrictive framework. \textbf{4)} It does not utilize any prior information at all, which is an excessively restrictive constraint. While collecting LR-HR image pairs are difficult, acquiring LR images only is relatively easy and feasible in many real-world scenarios. Since the internal-based SR methods generally show worse SR performance than the external-based models, it is desirable to exploit every available prior information as long as it stays in the unsupervised regime. 

To mitigate these limitations, we propose a simple baseline for unsupervised SR (SimUSR) that relaxes the ZSSR into a supervised setting. 
Instead of using a single image, our SimUSR make pseudo-pairs using multiple LR images. To correctly guide the model, we employ BM3D \cite{dabov2007image} to remove noises from the LR images when preparing the pairs. 
Though these are very simple corrections, they bring several benefits: our framework can now exploit every benefit of supervised learning. Thanks to this pseudo-supervision, ample prior information enables a model to reduce the performance gap between the unsupervised (only LR is available) and the supervised setting (HR is available). SimUSR can utilize recently developed network architectures and techniques that provide huge performance gains (\Fref{fig:teaser}). 
In addition, since the online training is not necessary, SimUSR can significantly reduce its runtime latency as well. The differences of the supervised SR, ZSSR, and SimUSR are summarized in \Fref{fig:overview}. 

We argue that our assumption is fairly practical while still remaining under the unsupervised learning setup; we only use LR images. Our approach is meaningful in that it investigates the blind spot of the field that should have been addressed but overlooked.

Our contributions are summarized as follows: 
\begin{enumerate}
    \item We propose a simple but strong baseline for unsupervised SR (SimUSR) 
    (NTIRE Challenge: 1st in PSNR, 2nd in SSIM, and 13th in LPIPS). 
    \item By casting the unsupervised SR into the supervised SR, SimUSR provides stable offline learning with a dramatically decreased latency at runtime. 
    \item We provide a comprehensive analysis on the effect of using various SR networks and techniques. By taking the state-of-the-art techniques and SR networks as our backbone, our method shows further enhancements.
\end{enumerate}

\section{Related work}
\noindent\textbf{Supervised Super-resolution.}
Recently, deep learning-based super-resolution models~\cite{ahn2018fast,dong2015image,kim2016accurate,lim2017enhanced,zhang2018image} have shown a dramatic leap over the traditional algorithms~\cite{timofte2014a+}. 
Most of the successful deep SR approaches fall into the supervised setting, where a network is trained on an external dataset having low- and high-resolution pairs. As long as the size of the dataset and the network capacity are large enough, it is well known that the supervised approach provides a better chance to enhance the SR performance~\cite{lim2017enhanced,zhang2018image}. However, it is also true that their performance and generalizability deteriorate dramatically when the dataset size is small and when there exists mismatch between training and testing environments~\cite{yoo2020rethinking, feng2019suppressing}.
To mitigate this issue, recent approaches focus on \textit{blind SR}, which assumes that there exist LR and HR pairs but with unknown degradation and downsample kernel~\cite{gu2019blind}.
%
Unlike the aboves, our proposed method can train a network even when there are no LR and HR pairs. 

\medskip
\noindent\textbf{Unpaired Super-resolution.}
A few recent works have addressed an unpaired SR task~\cite{fritsche2019frequency,lugmayr2019unsupervised,yuan2018unsupervised} that does not assume a paired setting. Since this setup does not require a full supervision, it is a more realistic scenario for many real-world applications. 
Most of the methods employ generative adversarial framework~\cite{goodfellow2014generative} so that a generator learns to map HR images into their distorted LR version. Using this generated pairs, they train an SR network in a supervised setting. 
However, in practice, there are cases where HR images are not even available, which requires a fully unsupervised SR. 

\medskip
\noindent\textbf{Unsupervised Super-resolution.}
Though the unpaired SR is sometimes considered as an unsupervised SR, we first clarify that unsupervised SR should strictly denote the task without any supervision neither paired images nor HR images. Under this definition, there are only a handful of studies \cite{ulyanov2018deep, heckel_deep_2018, shocher2018zero} and zero-shot super-resolution (ZSSR)~\cite{shocher2018zero} falls into this. ZSSR uses \textit{LR sons} that are downsampled images of the given LR test image (a.k.a \textit{LR father}). 
Using this pseudo pairs, they train the model in a supervised manner but only exploiting the internal statistics of the given test image. Because every procedure is performed at runtime, ZSSR suffered from high latency. To overcome this, Soh~\etal~\cite{soh2020meta} 
have proposed meta-transfer ZSSR (MZSR). 
They added a meta-transfer learning phase to exploit the information of the external dataset, which decreased the number of the steps required at runtime. Still, to quickly optimize the network, MZSR was limited to use a simple 8-layer network. Unlike the aforementioned methods, our SimUSR can benefit from the larger capacities of recently developed SR models and short latency at runtime by removing the online update phase, while remaining in the fully unsupervised regime in that it only utilizes the LR images. 
 
\section{Zero-shot super-resolution}
\label{sec:zssr}
The zero-shot super-resolution (ZSSR)~\cite{shocher2018zero} tackles the fully unsupervised SR task, where only low-resolution images ($\mathbf{I}_{LR}$) are available.
To do that, ZSSR performs both optimization and inference at runtime using a single test image (\Fref{fig:overview}).
During the online optimizing phase, they use an test input image ($\mathbf{I}_{LR}$) as \textit{LR father} ($\mathbf{I}_{LR}^{father}$) and generates \textit{LR son} ($\mathbf{I}_{LR}^{son}$) by downsampling LR father with an arbitrary kernel $k$. By doing so, 
they create \textit{pseudo-pair} $$(\mathbf{I}'_{LR}, \mathbf{I}'_{HR}) = (\mathbf{I}_{LR}^{son}, \mathbf{I}_{LR}^{father}),$$ where $\mathbf{I}_{LR}^{son} = \mathbf{I}_{LR}\downarrow_{s, k}$ and $\mathbf{I}_{LR}^{father} = \mathbf{I}_{LR}$. Here, $\downarrow_{s, k}$ denotes a downsampling operation with an arbitrary kernel $k$ and scale factor $s$.

With this pseudo-pair, optimizing a SR model now becomes a standard supervised setting. 
The core idea of ZSSR is to make the model learn internal image-specific statistics of a given test image during the online training. For inference, it generates final SR output ($\mathbf{I}_{SR}$) by feeding $\mathbf{I}_{LR}$ to the trained image-specific network. 



\begin{table*}[!t]
\centering
\caption{Quantitative comparison (PSNR/SSIM) on the bicubic SR (scale $\times$4) benchmark datasets. We boldface the best performance of both supervised SR and ours. }
\vspace{0.2em}
\setlength{\tabcolsep}{5.2pt}
\begin{tabular}{c|ccc|c|ccc}
\hline
\multirow{2}{*}{Dataset} & \multicolumn{3}{c|}{Supervised SR} & \multirow{2}{*}{ZSSR} & \multicolumn{3}{c}{SimUSR (Ours)} \\
\cline{2-4}\cline{6-8}
& CARN & RCAN & EDSR & & CARN & RCAN & EDSR \\\hline\hline
Set5     & 32.13/0.8937 & \textbf{32.63/0.9002} & 32.46/0.8968 & 31.13/0.8796 & 31.94/0.8908 & \textbf{32.40/0.8962} & 32.37/0.8955\\
Set14    & 28.60/0.7806 & \textbf{28.87/0.7889} & 28.80/0.7876 & 28.01/0.7651 & 28.44/0.7786 & \textbf{28.71/0.7860} & 28.70/0.7855\\
B100     & 27.58/0.7349 & \textbf{27.77/0.7436} & 27.71/0.7420 & 27.12/0.7211 & 27.49/0.7324 & \textbf{27.68/0.7394} & 27.66/0.7389\\
Urban100 & 26.07/0.7837 & \textbf{26.82/0.8087} & 26.64/0.8033 & 24.61/0.7282 & 25.70/0.7740 & \textbf{26.45/0.7986} & 26.31/0.7940\\
Manga109 & -            & \textbf{31.22/0.9173} & 31.02/0.9148 & 27.84/0.8657 & 30.03/0.9014 & \textbf{30.73/0.9124} & 30.59/0.9107\\\hline
\end{tabular}
\label{table:bicubic}
\end{table*}

\begin{table}[!t]
\centering
\caption{Quantitative comparison (PSNR) on SR (scale ×4) task with mixture of augmentation (MoA)~\cite{yoo2020rethinking}. We show the effect of MoA on our SimUSR and supervised SR (SSR) model. Note that SSR results are provided to show the improved upper limit again.}
\vspace{0.2em}
\setlength{\tabcolsep}{9pt}
\begin{tabular}{c|l|ccc}
\hline
Type & Model & Set14 & Urban & Manga \\\hline\hline
\multirow{2}{*}{SimUSR} 
                     & RCAN & \multirow{2}{*}{28.80} & \multirow{2}{*}{26.60} & \multirow{2}{*}{30.85} \\ 
                     & (+MoA) &  &  &  \\\hline
\multirow{2}{*}{SSR} 
                     & RCAN & \multirow{2}{*}{28.92} & \multirow{2}{*}{26.93} & \multirow{2}{*}{31.46} \\ 
                     & (+MoA) &  &  &  \\\hline
\end{tabular}
\label{table:moa}
\end{table}

\section{Our method}
\label{sec:method}
We introduce a simple baseline for a fully unsupervised super-resolution task (SimUSR). Similar to the ZSSR~\cite{shocher2018zero}, our method does not use any HR images for training the network. However, we slightly relax the constraint of ZSSR and assumes that it is relatively easy to collect the LR images, $\{\mathbf{I}_{LR_1},\dots,\mathbf{I}_{LR_N}\}$, where $N$ is the number of LR images. This allows our method to exploit multiple pseudo-pairs:
$$(\mathbf{I}'_{LR_k}, \mathbf{I}'_{HR_k}) = (\mathbf{I}_{LR_k}^{son}, \mathbf{I}_{LR_k}^{father}), \;\; \text{for}\;k = 1\dots N.$$ Here, we generate $\mathbf{I}_{LR}^{son}$ and $\mathbf{I}_{LR}^{father}$ with the same protocol that used in ZSSR. 

Though we now lose the generalizability over a single test image, compared to the cost of the relaxation, the benefits are very huge: we can fully enjoy the advantages of the supervised learning framework. More specifically, using these multiple pairs, we can now train a network offline and perform inference online as any supervised model usually does. Our method can be implemented by a simple modification of the supervised SR approach, it gives high flexibility and extensibility. For example, unlike the ZSSR and MZSR~\cite{soh2020meta}, which inevitably use shallow networks, we can use any off-the-shelf SR network and technique available, such as data augmentation~\cite{yoo2020rethinking} (\Sref{sec:bicubic}). In addition, since the runtime of our SimUSR only depends on the network's inference speed, this also gives a huge acceleration in terms of the runtime latency (\Sref{sec:time}).

\section{Experiments}
In this section, we describe our experimental settings and compare the performance of our method with the supervised SR models and the ZSSR~\cite{shocher2018zero}. In \Sref{sec:bicubic}, we analyze how much our SimUSR improves the performance over the ZSSR and how far we are left to reach the supervised performance. 
Then, in \Sref{sec:unknown}, we apply our method on the NTIRE 2020 SR dataset~\cite{NTIRE2020RWSRchallenge}.

\smallskip

\noindent\textbf{Baselines.}
We use ZSSR~\cite{shocher2018zero} as our major baseline method. However, since ZSSR and SimUSR are not designed to handle noisy cases, we attach BM3D~\cite{dabov2007image} as a pre-processing step. For our SimUSR, we use various models as our backbone network. We use three SR models: CARN~\cite{ahn2018fast}, RCAN~\cite{zhang2018image} and EDSR~\cite{lim2017enhanced}. Each of the model have different numbers of parameters from 1.1M to 43.2M (million).

\smallskip

\noindent\textbf{Dataset and evaluation.}
We use the DF2K~\cite{agustsson2017ntire,lim2017enhanced} dataset for the bicubic degradation SR task. However, unlike the Lim~\etal~\cite{lim2017enhanced}, we only use the LR images when we train the models.
For evaluation, we use Set5~\cite{set5}, Set14~\cite{yang2010image}, B100~\cite{b100}, Urban100~\cite{huang2015single}, and Manga109~\cite{matsui2017sketch} for bicubic SR task.
To evaluate our method on the real-world SR task, we use NTIRE 2020 dataset~\cite{NTIRE2020RWSRchallenge}.
This dataset is generated with unknown degradation operation to simulate the realistic image processing artifacts.
In addition, only non-paired LR and HR images are given so that the model should be trained via unsupervised setup.
Same as DF2K, we do not use any of HR images at the training phase. 
We use PSNR and SSIM to measure the performance. We calculate both metrics on RGB channels for the NTIRE dataset while only using the Y channel for the bicubic SR task. 

\begin{figure*}[!t]
\centering
\includegraphics[width=\linewidth]{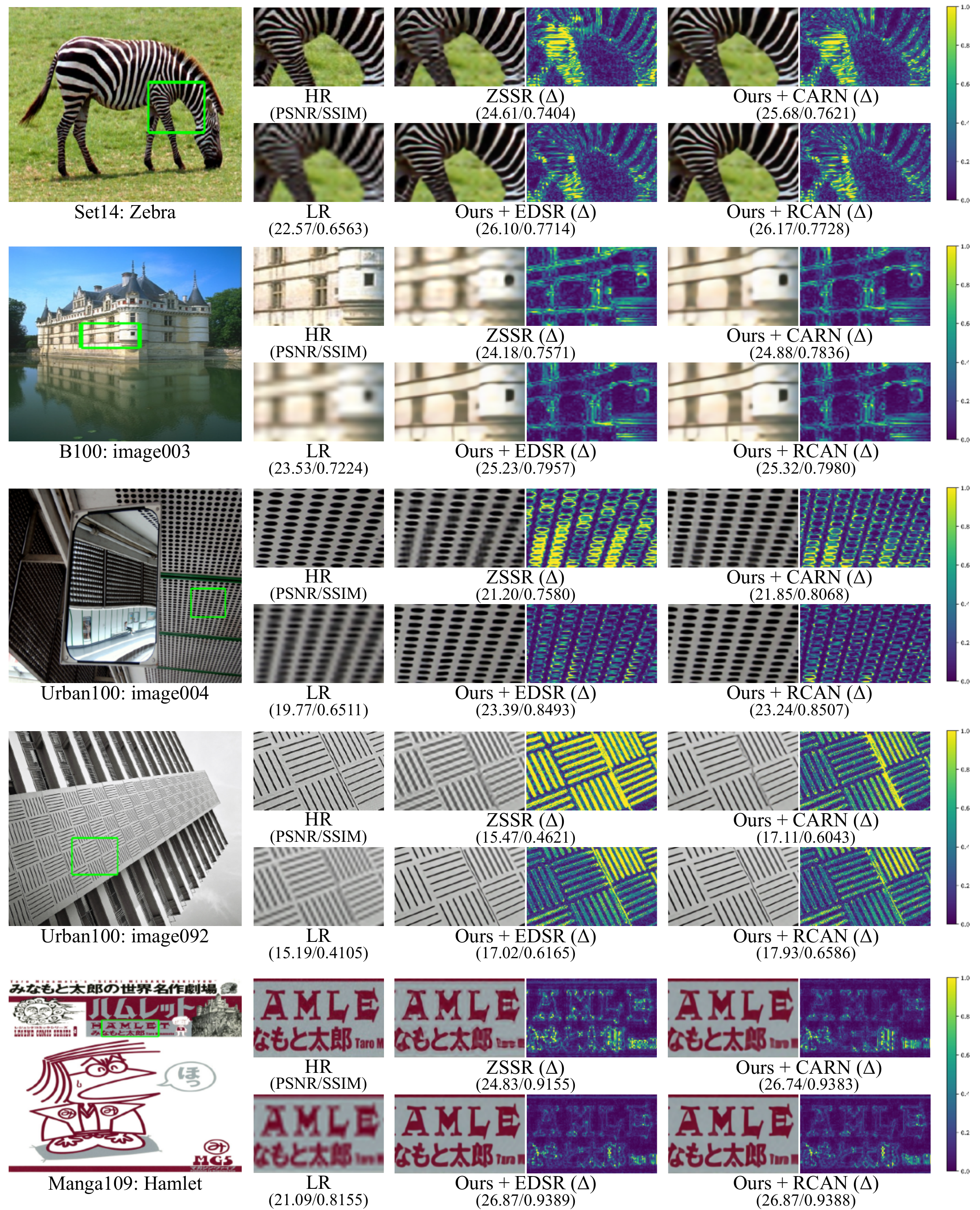}
\caption{Qualitative comparison of using our proposed method on the various benchmark datasets which are generated by the Bicubic downsample kernel. $\Delta$ is the absolute residual intensity map between the network output and the ground-truth HR image.}
\label{fig:comp_ntire_bix4}
\end{figure*}

\subsection{Bicubic super-resolution}
\label{sec:bicubic}
Here, we compare SimUSR with the ZSSR and the supervised SR models. Though the classical bicubic SR task is not our main task, it provides a testbed to analyze every model simultaneously. This also shows how much gap there exists between the supervised and unsupervised frameworks. For fair comparison, we report our performance using different SR networks as our backbone (CARN~\cite{ahn2018fast}, RCAN~\cite{zhang2018image}, and EDSR~\cite{lim2017enhanced}). The quantitative comparison on various benchmark dataset is shown in \Tref{table:bicubic}. Exploiting the additional LR images, our SimUSR shows large improvements over the ZSSR in every case. 

More interestingly, by exploiting the recent development of supervised SR techniques, such as data augmentation, SimUSR further reduces the gap toward the supervised learning models (\Tref{table:moa}). Note that, while the supervised models can use HR images as ground truth, SimUSR only uses LR images. Therefore, the model should generalize over the learned scale and pixel distributions. Toward this, we used mixture of augmentation (MoA)~\cite{yoo2020rethinking}, which is a recent data augmentation method for low-level vision task. MoA is known to not only improve the performance but also enhance the generalization power of the model. By employing the MoA, which ZSSR does not benefit from (results not shown), our performance again increases by 0.09 dB (Set14), 0.15 dB (Urban100), and 0.12 dB (Manga109), which are upto \textbf{3.63 dB} (Manga109) improvements over the ZSSR.  
Therefore, from now on, we use MoA with SimUSR by default unless it is specified. 

The qualitative results also shows the superior results of SimUSR over the ZSSR (\Fref{fig:comp_ntire_bix4}). In all the cases, SimUSR benefits from the increased performance by using external LR images. This tendency is clearly shown in the residual intensity map between the SR and HR image. For example, our method (with any backbone) successfully restores the replicating patterns (1st, 3rd, and 4th rows) while ZSSR has difficulty of recovering distortions. Note that ZSSR is supposed to better learn the internal statistics by repeatedly seeing the same LR image patches, which is in principle good at recovering replicating patterns. 
\begin{table}[!t]
\centering
\caption{Quantitative comparison (PSNR/SSIM) on the NTIRE 2020 dataset~\cite{NTIRE2020RWSRchallenge}. We analyze the effect of denoising (\textit{w/ BM3D}) and affine transformations (\textit{w/o Affine}). We also analyze the advantage of applying SimUSR.}
\vspace{0.2em}
\setlength{\tabcolsep}{7pt}
\begin{tabular}{c|cc|c}
\hline
\multirow{2}{*}{Method} & w/ & w/o & \multirow{2}{*}{PSNR / SSIM} \\
& BM3D & Affine \\\hline\hline
\multirow{3}{*}{ZSSR} &  & & 25.82 / 0.6898 \\
 & \cm & & 26.45 / 0.7320 \\
 & \cm & \cm & 26.55 / 0.7344 \\\hline
SimUSR+CARN & \cm & \cm & 27.19 / 0.7520 \\
SimUSR+RCAN & \cm & \cm & 27.24 / 0.7550 \\
SimUSR+EDSR & \cm & \cm & \textbf{27.28 / 0.7554} \\\hline
\end{tabular}
\label{table:real-world}
\end{table}

\begin{figure*}[!ht]
\centering
\includegraphics[width=\linewidth]{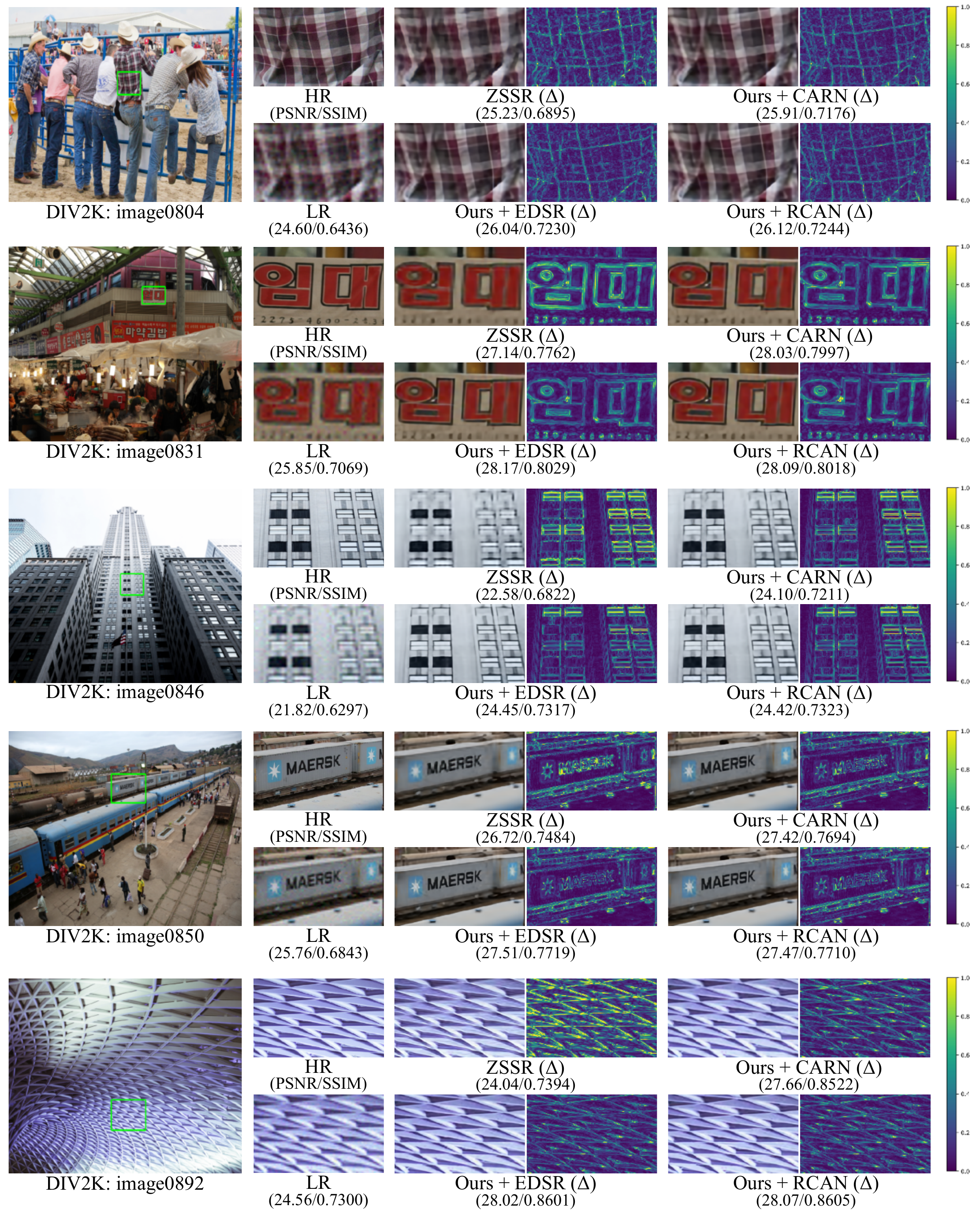}
\caption{Qualitative comparison of using our proposed method on the NTIRE 2020 dataset~\cite{NTIRE2020RWSRchallenge}. $\Delta$ is the absolute residual intensity map between the network output and the ground-truth HR image.}
\label{fig:comp_ntire}
\end{figure*}

\subsection{Real-world super-resolution}
\label{sec:unknown}

In this section, we compare ZSSR~\cite{shocher2018zero} and our method on the NTIRE 2020 dataset~\cite{NTIRE2020RWSRchallenge}. We found two observations that 1) ZSSR suffers from noise, and 2) the data augmentation methods, which are used in the original ZSSR, actually harm its SR performance (\Tref{table:real-world}). Based on this observation, we decided to attach BM3D~\cite{dabov2007image} before the ZSSR network optimization. For a fair comparison, we also use BM3D with our SimUSR. 
Regarding the data augmentation, we suspect that this is due to ZSSR network's small capacity and the severe spatial distortion by applying strong affine transformations~\cite{yoo2020rethinking}. 
 
By adding an ad-hoc denoiser (BM3D), ZSSR performance is dramatically improved by 0.63dB and 0.0422 in PSNR and SSIM, respectively. And by discarding affine augmentation, we can further enhance the ZSSR to achieve 26.55dB in PSNR (3rd row). With the same setting, our proposed SimUSR outperforms the ZSSR in a huge margin. For example, SimUSR with the lightweight SR network, CARN~\cite{ahn2018fast}, already boosts the SR performance of the ZSSR by 0.64dB and 0.0176 in PSNR and SSIM, respectively. 
Moreover, thanks to the high flexibility of our method, we can easily improve the performance by simply changing the backbone to any other SR network. For instance, we get another 0.09dB improvement in PSNR by just replacing a backbone network from CARN to RCAN~\cite{zhang2018image}. \Fref{fig:comp_ntire} shows the qualitative comparison between the ZSSR and our method with different backbone networks. Similar to the bicubic SR task, SimUSR (with any backbone) provides better restoration results across various cases.

\begin{table}[!t]
\centering
\caption{The number of the parameters and runtime comparison of 480$\times$320 LR images with scale factor $\times$4.}
\vspace{0.2em}
\setlength{\tabcolsep}{8pt}
\begin{tabular}{c|cccc}
\hline
\multirow{2}{*}{Method} & \multirow{2}{*}{ZSSR} & \multicolumn{3}{c}{SimUSR (Ours)} \\
\cline{3-5}
& & CARN & EDSR & RCAN  \\\hline\hline
\# Params. & 0.23M & 1.14M & 15.6M & 43.2M \\
Runtime & 300.83s & 0.12s & 1.93s & 1.07s \\ \hline
\end{tabular}
\label{table:time}
\end{table}

\begin{figure}[!ht]
\centering
\includegraphics[width=\linewidth]{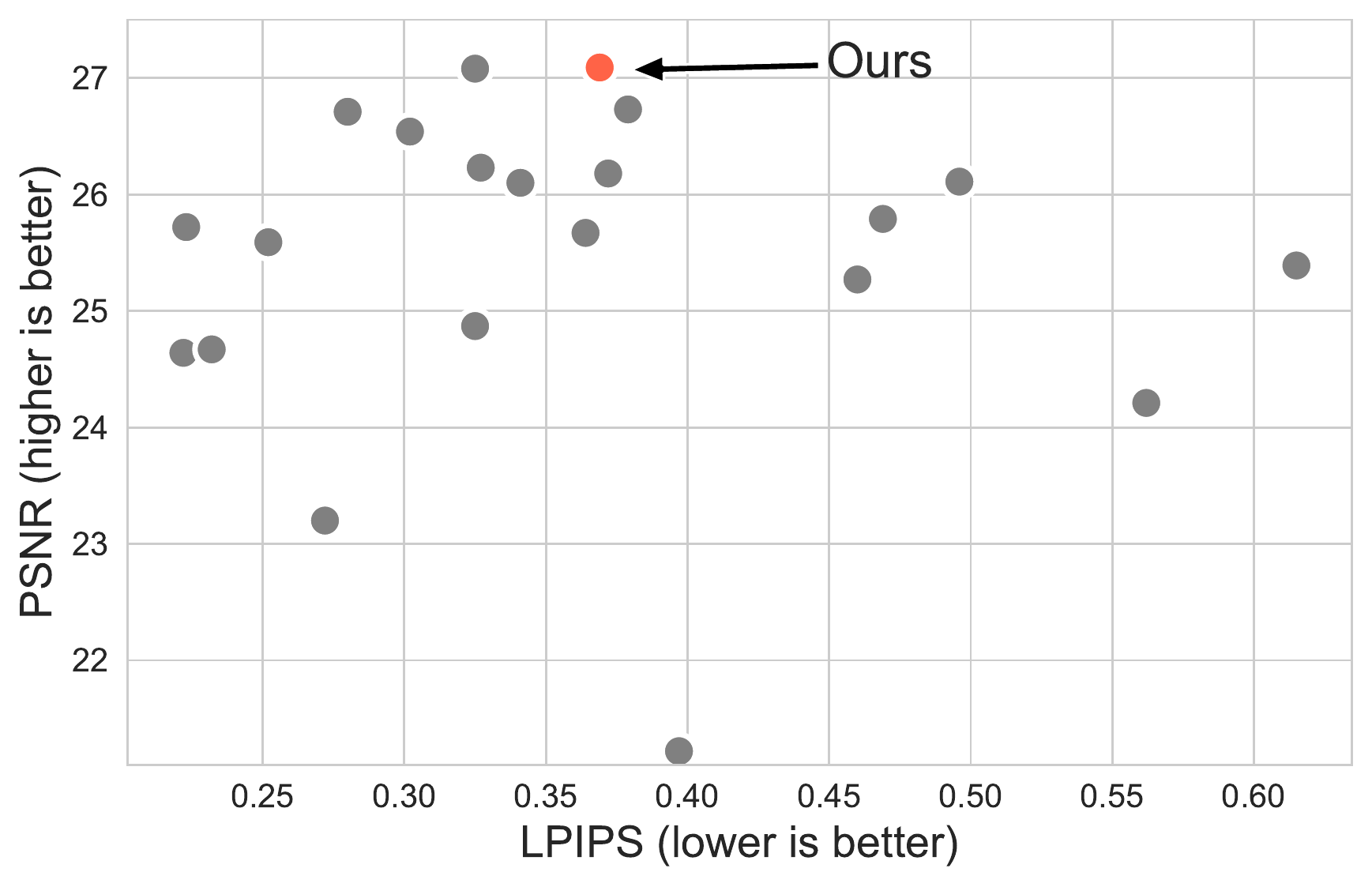}
\caption{Performance comparison of each entry in the NTIRE 2020 super-resolution challenge~\cite{NTIRE2020RWSRchallenge} (track one). Our proposed SimUSR is marked as a red circle. Our method achieves the best PSNR with a reasonable LPIPS (13th rank).}
\label{fig:challenge}
\end{figure}

\subsection{Execution time}
\label{sec:time}

In this section, we evaluate and compare the latency of ZSSR and our SimUSR (\Tref{table:time}). Note that we benchmark the runtime speed on the environment of NVIDIA TITAN X GPU by generating a 1080p SR image on scale factor $\times$4.
Although ZSSR has only 0.23M parameters, it requires a huge amount of runtime (300.83s) since it has to perform optimization and inference at runtime. 
In contrast, our proposed SimUSR only takes less than two seconds (1.93s) even if we use a heavy SR network (EDSR) as a backbone model. Comparing to the ZSSR, our method is at least 155 times faster than the ZSSR and if we use a lightweight SR network (CARN), 2,500 times faster (0.12s vs. 300.83s).

To embed the SR method to the real application, it is obvious that both SR performance and the latency are important aspects (\textit{e.g.} SR system for the streaming service).
However, the above analysis reflects that although ZSSR has nice properties, which does not need an HR image, applying it to the real application is challenging because of its high latency. On the other hand, our approach can meet the criteria that real applications demand (on both the performance and speed) by taking advantage of supervised SR. In addition, if necessary, we can further reduce the latency by replacing the backbone to more lightweight network thanks to the flexibility of our method.

\subsection{NTIRE 2020 super-resolution challenge}
\label{sec:ntire20}

This work is proposed to participate to the NTIRE 2020 super-resolution challenge~\cite{NTIRE2020RWSRchallenge}.
This challenge aims to develop an algorithm for the real-world SR task similar to the prior challenge in AIM 2019~\cite{AIM2019RWSRchallenge}.
However, unlike the previous challenge, there exist no LR and HR image pairs in the dataset akin to the setup that we experiment in \Sref{sec:unknown}. We submitted our SimUSR to the first track (image processing artifact), where the clean image is degraded by the unknown image artifact and downsample kernel. In this challenge, models are evaluated using the PSNR, SSIM and LPIPS~\cite{zhang2018unreasonable} metrics. 

Final result on the test dataset is shown in \Fref{fig:challenge} and \Tref{table:challenge}.
As shown in \Fref{fig:challenge}, our method achieves the best PSNR score among all entries with a reasonable LPIPS score. Note that since we directly optimize the network using pixel-based loss, the LPIPS score of our SimUSR is lower than the rank of PSNR. We also report the challenge result sorted on the PSNR (\Tref{table:challenge}). We get the best PSNR and second-best on SSIM with the 13th rank of LPIPS.

\section{Discussion}
In this section we discuss about the limitation of our method and the future direction. 

\medskip
\noindent\textbf{Limitation.}
Though the accessibility to multiple LR images is a mild and reasonable relaxation in many cases, there are still many applications and domains that cannot resort on such assumption where collecting the data is very expensive, \eg, medical imaging. 
In addition, SimUSR heavily relies on the generalizability of a model over different scales and pixel distributions, which can cause unexpected artifacts~\cite{yoo2020rethinking}. Because SimUSR uses bicubic downsampling to prepare the pseudo pairs, this may also cause an implicit bias in the SR model during the training. Last but not least, it is true that SimUSR is a basic approach that one would easily come up with but overlooked until now. We argue that it should be by no means a new state-of-the-art but serve as a reasonable baseline to beat in the future. 

\medskip
\noindent\textbf{Future work.}
We showed that our SimUSR framework is a strong baseline but it still has a plenty of room to improve its performance. For example, we used the BM3D as the pre-processing module for removing the noise. This pre-module can be replaced to more effective models~\cite{krull2019noise2void}. 

\begin{table}[!t]
\caption{Performance comparison of each entry in the NTIRE 2020 super-resolution challenge~\cite{NTIRE2020RWSRchallenge} (track one) sorted on the PSNR. The number in the parenthesis denotes the rank. Our proposed SimUSR is ranked the best in PSNR and second-best in SSIM.}
\vspace{0.2em}
\centering
\begin{tabular}{c|ccc}
\hline
Method       & PSNR $\uparrow$      & SSIM $\uparrow$     & LPIPS $\downarrow$     \\\hline\hline
\textbf{Ours}    & 27.09 (1) & 0.77 (2) & 0.369 (13) \\
Anonymous 1 & 27.08 (2) & 0.78 (1) & 0.325 (\hspace{0.5em}8)  \\
Anonymous 2 & 26.73 (3) & 0.75 (5) & 0.379 (15) \\
Anonymous 3 & 26.71 (4) & 0.76 (4) & 0.280 (\hspace{0.5em}6)  \\
Anonymous 4 & 26.54 (5) & 0.75 (8) & 0.302 (\hspace{0.5em}7)  \\
Anonymous 5 & 26.23 (6) & 0.75 (7) & 0.327 (10) \\\hline
\end{tabular}
\label{table:challenge}
\end{table}
\section{Conclusion}

We have introduced a simple but effective baseline for a fully unsupervised super-resolution task (SimUSR). 
we first clarify that unsupervised SR should strictly denote the task without any access to HR images. While complying with this definition, we assume that low resolution (LR) images are relatively easy to obtain in the real-world.  Exploiting multiple LR images, we generated a pseudo-pair dataset of LR images and their down-scaled version and use this to train a SR model. This simple conversion allows us to enjoy the advantages of supervised learning. We demonstrated that our SimUSR outperforms previous unsupervised SR method while having very short latency. Moreover, by integrating the recently developed SR architectures and techniques, we showed that SimUSR successfully close the performance gap between the unsupervised and the supervised SR methods. Though our approach is simple, we argue that accessibility to multiple LR images is a legitimate setting and SimUSR serves as a strong baseline of unsupervised SR in this regime, which should be investigated prior to considering other complicated methods.

\medskip

\noindent{\textbf{Acknowledgement.} This work was supported by NAVER Corporation and also by the National Research Foundation of Korea grant funded by the Korea government (MSIT) (no.NRF-2019R1A2C1006608)}

{\small
\bibliographystyle{ieee_fullname}
\bibliography{}
}

\end{document}